\documentclass[letterpaper, 10 pt, conference]{ieeeconf}
\IEEEoverridecommandlockouts    
\usepackage{graphics}           
\usepackage{times}              
\usepackage{amsmath}            
\usepackage{amssymb}            
\usepackage{graphicx}
\usepackage{algorithm}
\usepackage[noend]{algpseudocode}
\usepackage{booktabs}
\usepackage{color}
\usepackage[nocompress]{cite} 
\definecolor{instructioncolor}{rgb}{.5,.5,.5}

\usepackage[font=small]{caption}

\def\eqref#1{Eq.~(\ref{#1})}

\makeatletter
\usepackage{xspace}
\DeclareRobustCommand\onedot{\futurelet\@let@token\@onedot}
\def\@onedot{\ifx\@let@token.\else.\null\fi\xspace}

\makeatother

\usepackage{array}
\newcolumntype{L}[1]{>{\raggedright\let\newline\\\arraybackslash\hspace{0pt}}m{#1}}
\newcolumntype{C}[1]{>{\centering\let\newline\\\arraybackslash\hspace{0pt}}m{#1}}
\newcolumntype{R}[1]{>{\raggedleft\let\newline\\\arraybackslash\hspace{0pt}}m{#1}}

\usepackage{multirow}

\title{\LARGE \bf {Exploiting Motion Prior for Accurate Pose Estimation of Dashboard Cameras}}

\author{Yipeng Lu \and Yifan Zhao \and Haiping Wang \and Zhiwei Ruan \and Yuan Liu \and Zhen Dong \and Bisheng Yang 
  \thanks{
    Yipeng Lu, Haiping Wang, Zhen Dong and Bisheng Yang are with
    the Wuhan University 
    Yifan Zhao is with the China University of Geosciences,
    Zhiwei Ruan is with Didi Chuxing Technology Co.,
    Yuan Liu is with the University of Hong Kong. \textit{(Corresponding Author: Zhen Dong)}
  }%
  \thanks{
  }
}

\newcommand{\keywordss}[1]{\par\textbf{\textit{Index Terms---}}\textbf{#1}}

\begin{document}
\maketitle
\thispagestyle{empty}
\pagestyle{empty}

\begin{abstract}
  %



  

Dashboard cameras (dashcams) record millions of driving videos daily, offering a valuable potential data source for various applications, including driving map production and updates. A necessary step for utilizing these dashcam data involves the estimation of camera poses. However, the low-quality images captured by dashcams, characterized by motion blurs and dynamic objects, pose challenges for existing image-matching methods in accurately estimating camera poses. In this study, we propose a precise pose estimation method for dashcam images, leveraging the inherent camera motion {prior}. Typically, image sequences captured by dash cameras exhibit pronounced motion {prior}, such as forward movement or lateral turns, which serve as essential cues for correspondence estimation. Building upon this observation, we devise a pose regression module aimed at learning camera motion {prior}, subsequently integrating these {prior} into both correspondences and pose estimation processes. The experiment shows that, in real dashcams dataset, our method is {22\%} better than the baseline for pose estimation in AUC5\textdegree, and it can estimate poses for 19\% more images with less reprojection error in Structure from Motion (SfM).

\end{abstract}
\keywordss{ Dashboard camera, Motion pattern prior, Image matching, Pose estimation;}
\section{Introduction}
\label{sec:intro}

In recent decades, the widespread adoption of dashboard cameras has led to the recording of a significant amount of roadside videos every day. These recordings serve as vital resources for reconstructing traffic scenes and offer promising data sources for production and updating high-definition maps as well as spatial comprehension~\cite{zhanabatyrova2023automatic}. Leveraging these dashcam videos for the production and updating of high-definition maps holds the potential to significantly reduce mapping costs and enhance update frequency. However, the absence of integrated positioning sensors, such as GNSS receivers, IMUs, or LiDAR technology, presents a significant challenge to the direct utilization of dashcam videos for location-based applications~\cite{Chen2016vision}.

\begin{figure}
\begin{center}
\includegraphics[width=1\linewidth]{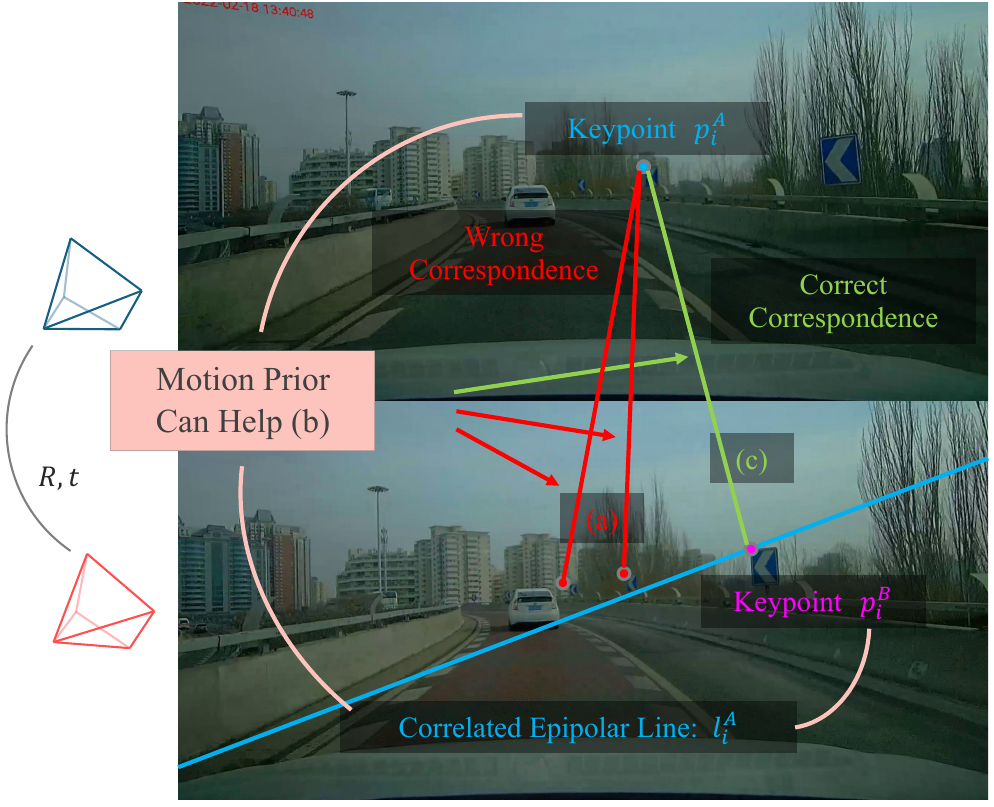}
\end{center}
\caption{{Dashcam images are often of low resolution with motion blur and dynamic objects, (a) which makes existing image matching methods struggle to estimate camera poses correctly. (b) In this paper, we propose to exploit the camera motion prior to restrict the correspondences to approximately conform to the coarsely estimated camera motions. (c) With the help of coarse camera motions, our method is able to accurately estimate correspondences, which thus results in accurate pose estimation. (updated the figure)}}
\label{fig:teaser}
\vspace{-25pt}
\end{figure}

To address this challenge, we propose a novel framework for pose estimation in dashcam imagery. This framework leverages inherent camera motion {prior} to enhance the accuracy of image matching. Since the dashcams are mounted on moving cars, these dashcams exhibit strong motion {prior} such as forward movement or lateral turns. These motion {prior} constrain the possible epipolar geometries of dashcam image pairs~\cite{chang2023structured}. 
As shown in Fig.~\ref{fig:teaser}, by approximating these camera motions, we are able to locate keypoints along coarse epipolar lines, thereby reducing the search space and improving the quality of correspondences. Consequently, the fundamental principle of our approach is to exploit these camera motion {prior} to refine the accuracy of pose estimation.

The first challenge lies in approximating these motion {prior}. Direct relative pose regression~\cite{kendall2015posenet} from the concatenated image features is able to capture these motion {prior}, but they often struggle to generalize to new data with unseen image {prior}. 
In our method, we regress the relative pose from feature correlations~\cite{sun2021loftr}, offering greater generalizability by relying on feature similarity rather than the features themselves.
The outcome of camera motion estimation comprises a coarse rotation and translation between image pairs.

The second challenge involves integrating the coarse estimates of camera motion to achieve precise correspondences. 
One straightforward method matches each keypoint along its estimated epipolar line, but inaccuracies in the regressed camera motion make it challenging to set an appropriate search region size.
Alternatively, our proposed matching method employs a soft constraint within the matcher, allowing it to autonomously learn an optimal search region. Building upon the SuperGlue framework~\cite{sarlin2020superglue}, our method enhances the matching process by incorporating encoding of the epipolar lines derived from the coarse camera motion, in addition to positional encoding of image coordinates. 

Furthermore, we incorporate the estimated coarse camera motion in the final pose estimation process to select a more reliable hypothesis. 
The motivation behind this integration is that, despite improved correspondence quality, incorrect matches persist and impede correct hypothesis selection. These errors typically cause the estimated poses to deviate from motion {prior}.
Therefore, the proposed method designs a scoring neural network to predict scores for different hypotheses incorporating the estimated coarse camera motions. This design significantly enhances the likelihood of selecting hypotheses that contribute to accurate pose estimation.

{The proposed method is trained solely on the KITTI dataset~\cite{Geiger2012CVPR} and is subsequently tested using both the NuScenes dataset~\cite{caesar2020nuscenes} and a Real Dashboard Camera dataset.
Sequences sourced from KITTI and NuScenes are of relatively high quality, whereas the self-collected sequence presents several challenges, including blurring, noise, and the presence of dynamic objects.}
Across all image sequences, our method consistently outperforms various baseline approaches in accurately estimating relative poses and producing {Structure-from-Motion (SfM)) reconstructions}.

In summary, we present the following contributions:
\begin{itemize}
    \item We proposed a method to regress relative poses from dense correspondences obtained through feature correlations, facilitating generalizability across different data sets. 
    \item We proposed a soft constraint mechanism in the matching, based on the coarse camera motion {prior}, enhancing the accuracy of correspondence selection.
    \item We proposed to incorporate the estimated coarse camera motion {prior} into the pose estimation process, employing a scoring neural network to improve the selection of hypotheses for accurate pose estimation. 
\end{itemize}

\section{Related Work}
\label{sec:related}



\begin{figure*}
\begin{center}
\includegraphics[width=1\linewidth]{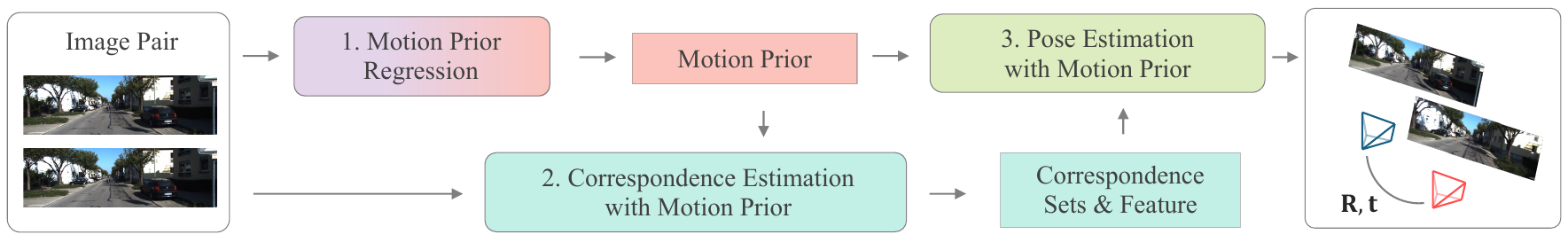}
\end{center}
\vspace{-10pt}
   \caption{Overview of the proposed method. 
   {The motion prior regression module (1) first estimates a coarse relative pose of the input image pair by leveraging the motion prior. Then, the estimated coarse relative pose is incorporated into correspondence estimation (2) and pose estimation (3) to obtain a more accurate camera pose.(updated the figure)}
   }
\label{fig:pipeline}
\vspace{-20pt}
\end{figure*}

\subsection{Relative Pose Regression}
Instead of solving relative camera poses from established correspondences, a number of recent works
~\cite{parameshwara2022diffposenet,cai2021extreme,liu2022gen6d} 
directly regress relative camera poses of a given pair of images from their features.
These methods leverage the strong motion {prior} exhibited in daily life image capturing~\cite{cavalli2022nefsac}, which constrains the motion of the camera within a predictable range or trajectory even in unseen scenarios~\cite{arnold2022map}.  
The dashboard cameras typically show strong {motion prior} such as moving forward or turning left/right. 
{Thus we propose to learn to encode such motion prior with a neural network similar to GRelPose~\cite{khatib2022grelpose}, but focusing solely on geometric relationships rather than incorporating additional features.}
We then use the encoded {motion prior} to improve correspondence estimation and pose estimation, leading to more accurate relative camera pose estimation.
 
\subsection{Correspondence estimation}
Traditional methods estimate correspondences between hand-crafted
~\cite{rublee2011orb,bay2006surf}
or learning-based
~\cite{liu2019gift,revaud2019r2d2,detone2018superpoint,9636619}
local features by nearest neighbour search with mutual check or ratio check~\cite{lowe2004distinctive}. 
In recent developments, detector-based
~\cite{sarlin2020superglue, lindenberger2306lightglue} 
matching methods such as SuperGlue~\cite{sarlin2020superglue} have achieved significant improvements. 
Recently, detector-free matches
~\cite{sun2021loftr,chen2022aspanformer, edstedt2023dkm, edstedt2024roma} 
enhance the input features with attention-based GNNs and match similar enhanced features to correspondences, which improves the correspondence quality. 
Such methods obviate the necessity for keypoint detection and extraction, performing well in texture-less environments, such as indoor scenes. However, the absence of stable keypoints also makes it challenging to handle downstream applications like SfM.
Other methods incorporate external information to improve performance{~\cite{10244083,9353970,9801615, yifan2022input}.}
{However, dashboard images often include compression artefacts of the textureless regions, trails of motion blur, repetitive patterns, and dynamic objects, resulting in indiscriminative and ambiguous local features~\cite{mur2015orb}, where the aforementioned methods struggle to find reliable correspondences by solely relying on feature similarity. }

\subsection{Model scoring in pose estimation}
Given estimated correspondences, the RANSAC~\cite{fischler1981random} paradigm is widely adopted to recover the relative camera poses of the image pair. 
Traditional methods
~\cite{barath2021efficient,barath2021marginalizing,7139705} 
propose model scoring functions based on inlier counting or well-designed maximum likelihood procedures, which are sensitive to the inlier-outlier threshold setting or inlier-outlier distribution.
Recently, MQ-Net~\cite{barath2022learning} learns to score the model from point-to-model residual distribution and achieves impressive accuracy. 
However, in real-world scenarios of dashboard images, outlier correspondences tend to form spatially coherent structures due to repetitive patterns or dynamic objects. The aforementioned methods are prone to trap plausible geometric models from these coherent outliers. We additionally incorporate the estimated motion {prior} into a model scoring network to resist plausible models.

\section{Methodology}
\label{sec:main}


Given two images $A$ and $B$ with the known intrinsic matrix {$\mathbf{K}$}, our goal is to recover the relative camera pose {$(\mathbf{R,t})$} between these two images. Our pipeline is illustrated in Fig.~\ref{fig:pipeline}. 
{
In the following sections, we first provide a coarse estimate of the relative pose in Section III-A.
Next, we explain how to use this initial estimate to find correspondences between the two images in Section III-B, 
and how to refine the pose estimation accuracy in Section III-C.}

\subsection{Motion {prior} regression}
\label{sec:pose_reg}
\begin{figure}
\begin{center}
\includegraphics[width=1\linewidth]{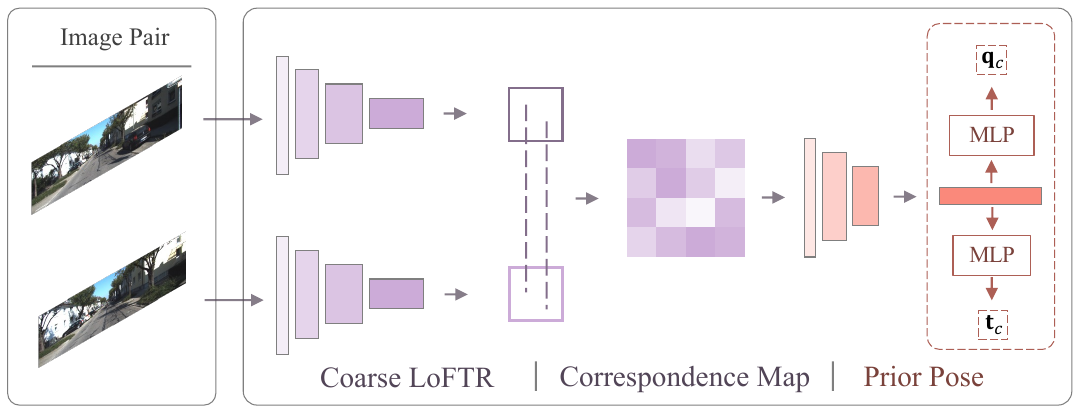}
\end{center}
\vspace{-10pt}
   \caption{{The motion {prior} regression module estimates a coarse camera pose for the input image pair.(updated the figure)}}
\label{fig:pose_reg_arc}
\vspace{-20pt}
\end{figure}

As the dashcam image sequences often show strong camera motion {prior} that can be used for pose regression, we introduce a neural network to regress the relative camera motion in this part. 
Fig.~\ref{fig:pose_reg_arc} shows the overall architecture of the motion {prior} regression module. The module first extracts dense correspondences by correlating every feature vector of image $A$ with those of image $B$ and then regresses the rotation and translation from the extracted correspondence.

\textbf{Coarse feature extraction}.
The module begins with extracting features at 1/8 of the original image size from both images. In this component, it adopts the pre-trained coarse feature extraction layers of LoFTR~\cite{sun2021loftr}.

\textbf{Coarse correspondence estimation}.
Subsequently, for each feature vector $f_c^A(u,v)$ extracted from the feature map $F_c^A$ of image $A$, the module computes its correlation $r(u,v)$ with each feature vector $f_c^B(u',v')$  extracted from the feature map $F_c^B$ of image $B$. $(u,v)$ denotes the coordinate of the feature map. $Cor(\cdot)$ denotes the correlation between the two feature vectors{as following:}
\begin{equation}
    Cor[f_c^A(u,v), f_c^B(u',v')] = f_c^A(u,v) \cdot f_c^B(u',v')
\end{equation}
\begin{equation}
    r(u, v) = max(Cor[f_c^A(u,v), f_c^B(u',v')]), \forall f_c^B(u',v') \in F_c^B
\end{equation}

Following this, for each feature $f_c^A(u,v)$ , the $f_c^B(u',v')$  from $F_c^B$ with the maxium correlation is selected as the correspondence $(u, v) - (u', v')$. For every pair of correspondences, their coordinates and correlations are concatenated together as the correspondence map $S_c$ {as following:}
\begin{equation}    
    S_c(u,v) = [u, v, u', v', r(u, v)]
\end{equation}

\textbf{Coarse prior pose regression}.
Consequently, for every feature position derived from $F_c^A$, a 5-dimensional vector $S(u,v)$ can be constructed. The dimensions of the correspondence map $S$ are $(\frac{1}{8}H,\frac{1}{8}W,5)$.

Then we employ a ResNet-like architecture on the correspondence map $S$ to perform regression, estimating both coarse rotation {$\mathbf{R}_c$} as a quaternion {$\mathbf{q}_c$} and coarse translation {$\mathbf{t}_c$} as a unit vector. To train the regressor, we utilize the L1 loss function measuring the disparity between the predicted rotations and translations and the ground-truth values.

\subsection{Correspondence estimation with motion {prior}}
\label{sec:corr_est}

\begin{figure*}
\begin{center}
\includegraphics[width=1\linewidth]{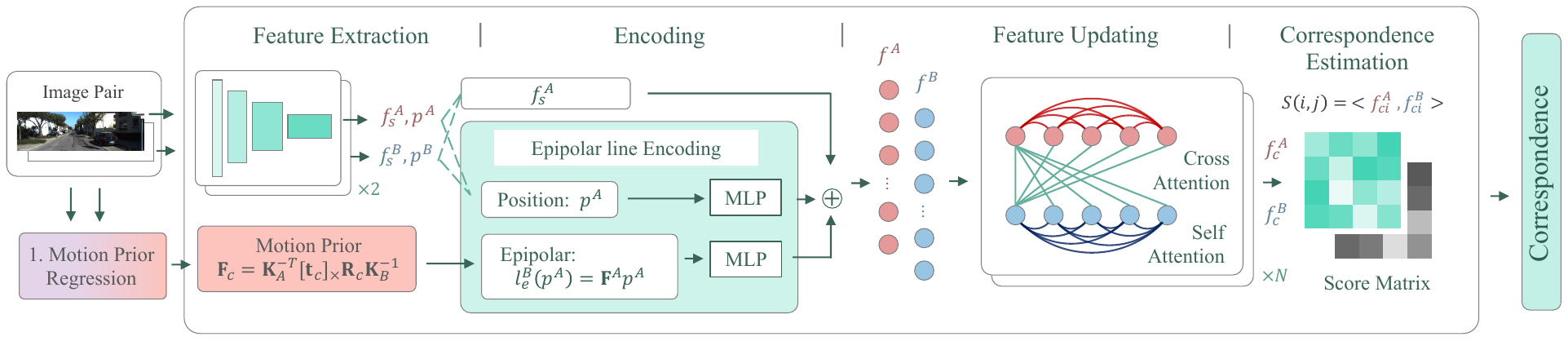}
\end{center}
\vspace{-10pt}
\caption{{Correspondence estimation with motion prior. 
The proposed Epipolar Line Encoding encodes the coarse camera pose to features extracted on the image pair. Then the features are fed to several interleaved cross- and self- attention layers for feature updating and correspondence estimation.(updated the figure)}}
\label{fig:correspondence_estimation}
\vspace{-20pt}
\end{figure*}

In this section, we aim to establish correspondence sets between the image pair $(A,B)$. Fig.~\ref{fig:correspondence_estimation} shows the overall architecture.

{The coarse relative pose $(\mathbf{R}_c, \mathbf{t}_c)$ estimated in Section \ref{sec:pose_reg} can serve as a prior}, reducing the search regions for identifying correct correspondences.
Hence, we integrate coarse relative camera poses into a transformer-based matcher~\cite{sarlin2020superglue} using epipolar line encoding along with keypoints position encoding.

{Given a set of keypoints' features ${f_s^A}, {f_s^B}$ and their positions $p^A, p^B$ from images $A$ and $B$}, along with the prior coarse pose {$(\mathbf{R}_c, \mathbf{t}_c)$} regressed beforehand, the proposed matcher returns the correspondence set $\mathbf{C}_{A,B}$ for these keypoints.

\textbf{Feature extraction}.
For each pair of images $A$ and $B$, the module initially extracts a set of keypoints, including their features $f_s^A, f_s^B$, and positions $p^A, p^B$ in the pixel coordinate by SuperPoint~\cite{detone2018superpoint}.

\textbf{Keypoints position encoding}.
The transformer module requires position encoding to distinguish different features. Following SuperGlue~\cite{sarlin2020superglue}, the keypoint positions in pixel coordinates are initially transformed into the camera coordinate using intrinsic parameters {$\mathbf{K}_A, \mathbf{K}_B$}. Subsequently, a simple MLP $\phi_p(\cdot)$ is utilized to map them to a higher dimensionality to match the feature dimensions.

\textbf{Epipolar line encoding}.
Similarly, in the epipolar encoding stage, another MLP is employed to encode the epipolar line to align with the feature dimension.
Given the coarse relative pose {$(\mathbf{R}_c, \mathbf{t}_c)$} of image $A, B$, with their intrinsic matrix {$\mathbf{K}_A, \mathbf{K}_B$}, it is able to compute the corresponding fundamental matrix {$\mathbf{F}^A, \mathbf{F}^B$}.
{Subsequently, for the keypoints  $p^A$}, it is able to compute its corresponding epipolar line in image $B$ in the camera coordinate{as following:
\begin{equation}
\left\{
\begin{aligned}
    &l_e^B = \mathbf{F}^A p^A = (\mathbf{K}_B^{-1})^T [t_c]_{\times}R_c \mathbf{K}_A^{-1} p^A \\
    &l_e^A = \mathbf{F}^B p^B = (\mathbf{K}_A^{-1})^T [t_c]_{\times}R_c \mathbf{K}_B^{-1} p^B 
\end{aligned}
\right.
\end{equation}}
$l_e^B$ denotes the corresponding epipolar line of $p^A$ in image $B$, whereas $l_e^A$ denotes the corresponding epipolar line of $p^B$ in image $A$.
Following this, the module normalizes $l_e^A, l_e^B$ to unit vectors and then applies an MLP { $\phi_e(\cdot)$ } to encode the epipolar line, which serves as the encoding feature of the epipolar line.

\textbf{Feature updating}.
With the previously extracted features $f_s^A, f_s^B$, 
{point coordinate encoded features $f_p^A, f_p^B$ and epipolar line encoded features $f_e^A, f_e^B$}, the module aggregates them by simple addition to form the input feature. 
Subsequently, these merged features undergo a sequence of self-cross attention layers, enabling information exchange within and between the images. 
Following several layers of self-cross attention, the features of keypoints are updated for both images.


\textbf{Matching score prediction}. 
Afterwards, a scoring matrix $S \in \mathbb{R}^{N_{A}\times N_{B}}$ is formulated using the updated features. Here,  $S(i,j) = \langle f^A_{ci}, f^B_{cj} \rangle$ denotes the inner product between the updated features {$f^A_{c}, f^B_{c}$}, where  $N_A, N_B$ {represent} the numbers of keypoints on image $A$ and image $B$ and {$f^A_{ci} \in f^A_{c}$, $f^B_{ci} \in f^B_{c}$}. The elements $S(i, j)$ in the score matrix represent the matching confidence of keypoints {$p^A and p^B$ , where $p_i^A \in p^A$ and $p_j^B \in p^B$.}

\textbf{Correspondence estimation}. 
The issue can be reframed as an optimal transport problem. In this formulation, each keypoint in image $A$ is allocated to a keypoint in image $B$ based on a cost matrix, derived from the previously calculated score matrix $S$. The entropy regularization of $S$ facilitates a softer assignment of correspondences. This problem can be effectively addressed using the Sinkhorn algorithm, which iteratively normalizes the rows and columns of the cost matrix, thereby converging towards a soft assignment. Consequently, the correspondences are obtained.

\subsection{Pose estimation with motion {prior}}
\label{sec:pose_est}

Given the correspondences estimated in the preceding section, the primary objective of this section is to determine the relative pose based on these correspondences. Initially, a straightforward method is employing RANSAC and the 5-point algorithm for pose estimation. 

The typical RANSAC iteratively selects hypotheses consisting of five correspondences until reaching the maximum iteration. Each hypothesis is then used to compute the essential matrix {$\mathbf{E}$}, assessing the Sampson distance and classifying it into inliers or outliers. The hypothesis with the most inliers is chosen, and the rotation and translation are determined accordingly.

\begin{figure}
\begin{center}
\includegraphics[width=1\linewidth]{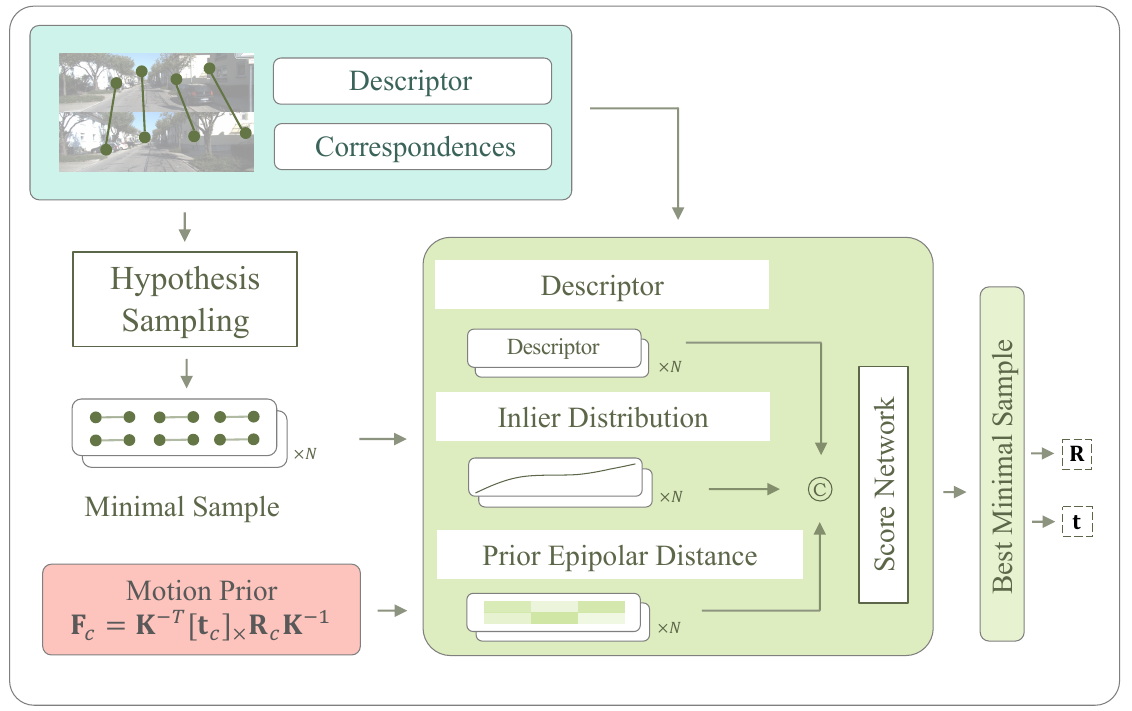}
\end{center}
\vspace{-10pt}
\caption{{Pose estimation with motion prior. The estimated motion prior and inlier distribution are utilized when scoring the camera pose hypotheses in RANSAC.(updated the figure)}}
\label{fig:ransac}
\vspace{-20pt}
\end{figure}

In this section, we follow the overall RANSAC framework but employ an improved method for hypothesis scoring. Instead of relying solely on inlier count, our scoring mechanism involves the utilization of a neural network that integrates {motion prior} and {distribution of inliers}. 
Fig.~\ref{fig:ransac} shows the overall architecture.

\textbf{Hypothesis sampling}. 
In our implementation, we first sample {$N$} hypotheses from the estimated correspondences, followed by the selection of the top 100 hypotheses based on their inlier count. Here, a hypothesis denotes a grouping of correspondences. 
Traditionally, a minimum of 5 correspondences suffices for estimating relative pose. However, this minimal setup might result in multiple solutions. To alleviate this uncertainty, we randomly sampling 6 correspondences for each hypothesis. Following this, the inlier count is computed for each hypothesis using the remaining correspondences, and the top 100 candidates are then determined based on the inlier count. Inliers and outliers are delineated by Sampson distance threshold $d_\epsilon$. Sampson distance $d(p_i^A, p_i^B)$ can be computed as:
\begin{equation}
    d(p_i^A, p_i^B) = \frac{({p_i^B}^T \mathbf{F}_i {p_i^A})^2}{(\mathbf{F}_i {p_i^A})^2_x+(\mathbf{F}_i {p_i^A})^2_y+({p_i^B}^T \mathbf{F}_i)^2_x+({p_i^B}^T \mathbf{F}_i)^2_y}
\end{equation}
where $\mathbf{F}_i$ is the fundamental matrix and $\mathbf{F}_i=\mathbf{K}_B^{-T} \mathbf{E}_i \mathbf{K}_A^{-1}$
These candidate hypotheses are then evaluated using a scoring neural network to determine the most accurate {one}. 

\textbf{Hypothesis scoring: motion {prior} perception}.\label{sec:pose_est:coarse_pose}

The coarse relative pose, established in Sec.~\ref{sec:pose_reg}, defines the general camera motion {prior} represented by $\mathbf{R}_i$ and $\mathbf{t}_i$. For each hypothesis, we calculate the Sampson distance of the 6 seed correspondences under the {prior pose fundamental matrix} $\mathbf{F}_i=\mathbf{K}_B^{-T} [\mathbf{t}_i]_{\times} \mathbf{R}_i \mathbf{K}_A^{-1}$ and feed this feature vector into the hypothesis scoring network to approximate the regressed relative pose.

\textbf{Hypothesis scoring: inlier distribution perception}.\label{sec:pose_est:inlier_dis}
Compared with the inlier number, the cumulative distribution of inlier provides more information on hypothesis quality~\cite{barath2022learning}.
Inliers corresponding to each hypothesis are determined by applying {a Sampson distance threshold $d_\epsilon = 12.6$. Subsequently, $d_\epsilon$ is divided into $n_b=64$ bins where each bin denotes $0.2$}, and the number of inliers within each bin is tabulated. Each bin represents a Sampson distance $\delta_i = \frac{i}{n_b} d_\epsilon$, uniformly distributed between $0$ to $d_\epsilon$.  For each bin, its value denotes the ratio of inliers with a Sampson distance below $\delta_i$  to all correspondences{, given by}
\begin{equation}
    b_k = \frac{n_{i}(k)}{n_{c}}=\frac{I(d_i < \delta_k)}{n_c}, \ k \in [0, n_b]
\end{equation}
Here, $b_k$ signifies the $k$-th bin inlier ratio, $n_i(k)$ represents the number of inliers in the $k$-th  bin, $n_c$ stands for the total number of correspondences, 
$I(\cdot)$ {indicates} the count that satisfies the condition, and  $I(d_i < \delta_k)$ indicates the inlier count with a Sampson distance below $\delta_k$.

\textbf{Hypothesis scoring: network implementation}. 
As shown in {Fig.~\ref{fig:ransac}}, the scoring network receives inputs from the inlier distribution, comprising $m=64$ bins, the 6 epipolar distances under the prior coarse relative pose, and the descriptors of hypothesis from the matcher. Subsequently, these inputs are processed by {MLPs with ReLU activation and batch normalization layers}. The {batch normalization} layers standardize features across the top 100 hypotheses. Finally, the scoring network produces a score for each hypothesis, and the hypothesis with the highest score is chosen as the output hypothesis, which is subsequently decomposed to derive the final $\mathbf{R}_i, \mathbf{t}_i$.

\textbf{Hypothesis scoring: loss function}. 
The score network employs binary cross-entropy loss. 
{The network outputs scores for each hypothesis and the estimated pose is utilized for label calculation. }
Labels are determined based on the angular errors in rotation \( R_{err} \) and translation \( t_{err} \) between the estimated pose and ground truth pose. { To ensure than $R_{err}$ and  $t_{err}$ contribute equally, we average the two errors and map through a continuous linear function from the range $[0^{\circ}, 20^{\circ}] \rightarrow [1, 0]$ to obtain the network labels.} Errors exceeding $ 20^{\circ}$ are considered negative labels.

\section{Experimental Evaluation}
\label{sec:exp}

%

%
%
%
%

\subsection{Experimental Setup}




\paragraph{Datasets}
{The method is evaluated using the KITTI~\cite{Geiger2012CVPR} dataset, the NuScenes dataset~\cite{caesar2020nuscenes} and the RealDashCam(RDC) dataset. }

{For the KITTI dataset, we randomly selected frame intervals ranging from 5 to 13 to generate image pairs, resulting in 62,833 pairs for training, 800 pairs for validation, and 2,347 pairs for testing.}

{For the NuScenes dataset, we selected 36 scenes to generate image pairs solely for testing. Images were randomly chosen with frame intervals ranging from 20 to 30, resulting in a total of 14,922 test image pairs.}

\begin{figure}
\begin{center}
    \includegraphics[width=1\linewidth]{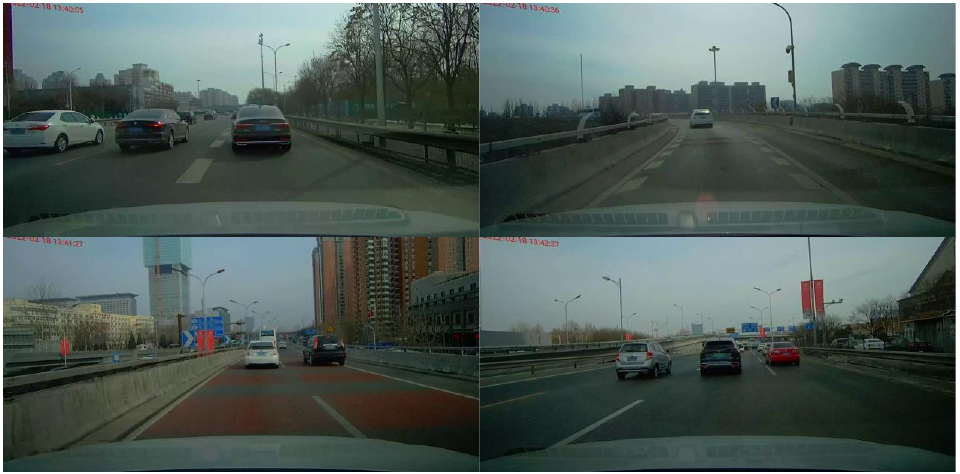}
\end{center}
\vspace{-10pt}
\caption{Example data and trajectories of RDC dataset}
\label{fig:rdc_data_traj}
\vspace{-20pt}
\end{figure}

{For the RealDashCam (RDC) dataset, it was collected in Beijing by us.}
It comprises totalling 1,348 images at a resolution of $2284 \times 1123$. This dataset is solely utilized for evaluation. These images exhibit low quality and contain multiple dynamic objects and compression artefacts, presenting significant challenges for accurate pose estimation.  The image examples are illustrated in {Fig.~\ref{fig:rdc_data_traj}. Image pairs from the RDC dataset are randomly selected with frame intervals ranging from 15 to 25, resulting in 2328 test image pairs.} The ground truth poses are derived from GNSS trajectories and refined through {SfM}.

\begin{table*}[h]
\centering
\caption{{Quantitative results on the KITTI \& NuScene \& RDC dataset}}
\label{tab:auc_result}
\resizebox{0.99\linewidth}{!}
{
\begin{tabular}{clccccccccccc}
\toprule
   & \multirow{2}{*}{Method} & \multicolumn{3}{c}{AUC of KITTI} & & \multicolumn{3}{c}{{AUC of NuScene}} & & \multicolumn{3}{c}{AUC of RDC} \\
  \cmidrule{3-5} \cmidrule{7-9} \cmidrule{11-13} &                   & $5^{\circ}$  & $10^{\circ}$  & $20^{\circ}$  & & $5^{\circ}$ & $10^{\circ}$  & $20^{\circ}$ &  & $5^{\circ}$ & $10^{\circ}$  & $20^{\circ}$    \\
\midrule
\multirow{4}{*}{Matcher}   
        & SuperGlue~\cite{sarlin2020superglue}      & {0.7454} & {0.8654} & {0.9303} & & {0.6316} & {0.7379} & {0.7991} & & {0.4486} & {0.6008} & {0.6976}  \\
        & LoFTR~\cite{sun2021loftr}                 & {0.7775} & {0.8812} & {0.9380} & & {0.6409} & {0.7406} & {0.7988} & & {0.3769} & {0.5117} & {0.6120}  \\
        & AspanFormer~\cite{chen2022aspanformer}    & {0.7759} & {0.8781} & {0.9346} & & {0.6749} & {0.7719} & {0.8258} & & {0.3084} & {0.4567} & {0.5767}  \\
        & DKM~\cite{edstedt2023dkm}                 & {0.7357} & {0.8585} & {0.9247} & & {0.6789} & {0.7693} & {0.8197} & & {0.3342} & {0.4497} & {0.5416}  \\
        & RoMa~\cite{edstedt2024roma}               & {0.7598} & {0.8728} & {0.9337} & & {0.7014} & {0.7839} & {0.7839} & & {0.4614} & {0.5656} & {0.6396}  \\
\midrule
\multirow{2}{*}{Estimator} 
        & NeFSAC~\cite{cavalli2022nefsac}           & {0.7632}   & {0.8741}  & {0.9345} & & {0.6436} & {0.7413} & {0.7994} & & {0.4050} & {0.5626} & {0.6699}  \\
        & MQNet~\cite{barath2022learning}           & {0.7608}   & {0.8623}  & {0.9315} & & {0.6279} & {0.7161} & {0.7647} & & {0.4179} & {0.5451} & {0.6230}  \\
\midrule
        & Ours                                      & {\textbf{0.7998}}   &  {\textbf{0.8906}}   & {\textbf{0.9404}} & & {\textbf{0.7135}}   &  {\textbf{0.7862}}   & {\textbf{0.8251}} & & {\textbf{0.5731}}  & {\textbf{0.6909}}  & {\textbf{0.7679}} \\
\bottomrule
\end{tabular}
}
\vspace{-10pt}
\end{table*}

\paragraph{Baselines}
In this study, two types of matching methods were employed: detector-based methods and detector-free methods. The representative of the detector-based method is SuperGlue~\cite{sarlin2020superglue}.
The representative of the detector-free method is LoFTR~\cite{sun2021loftr},Aspanformer~\cite{chen2022aspanformer}, DKM~\cite{edstedt2023dkm}, and RoMA~\cite{edstedt2024roma}. All of these matching methods utilize RANSAC for pose estimation.

For the pose estimation method, two learning-based approaches were employed: NefSAC~\cite{cavalli2022nefsac} and MQNet~\cite{barath2022learning}, both of which are learning-based and can capture certain regularities of correct sampling, thereby enhancing the accuracy of camera pose estimation. {These pose estimation baselines take SuperGlue~\cite{sarlin2020superglue} correspondence as input.}

\paragraph{Metrics}
In accordance with prior methodologies~\cite{sarlin2020superglue,zhang2019learning}, performance assessment is conducted through Area-Under-Curve (AUC) metrics derived from pose accuracy curves.  
For each estimated relative pose angular disparities between rotations and normalized translation vectors are computed in comparison to the corresponding ground-truth pose. Subsequently, the pose error is determined as the larger value between the angular errors in rotation and translation. 
Specifically, we report the AUC values corresponding to angular errors less than 5$^\circ$, 10$^\circ$, and 20$^\circ$ across all {experiment.}

\begin{figure*}
\begin{center}
    \includegraphics[width=1\linewidth]{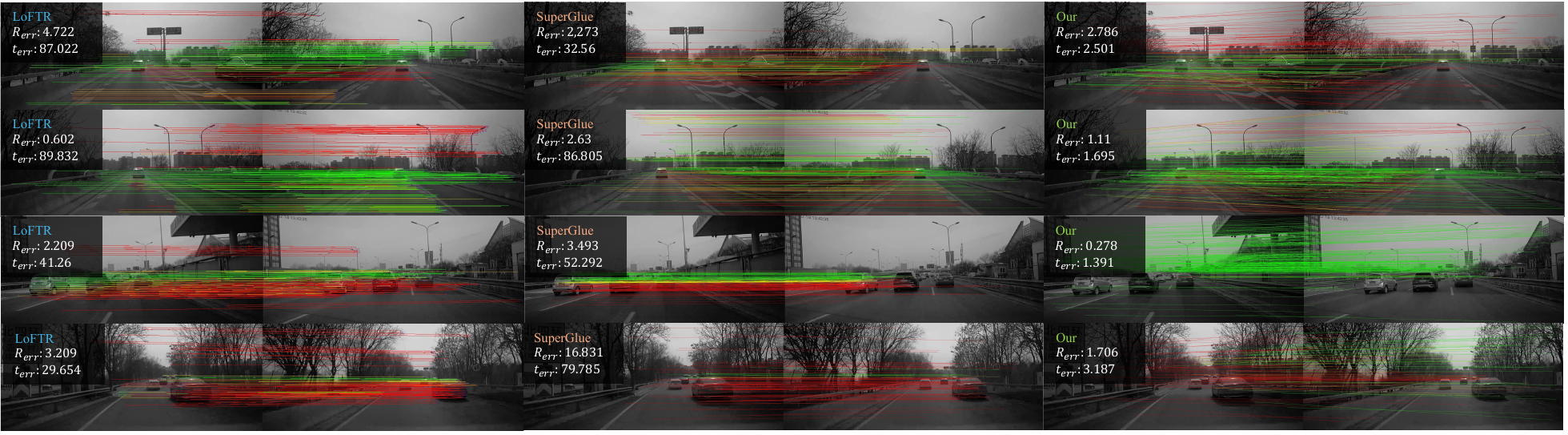}
\end{center}
\vspace{-10pt}
\caption{\textbf{Example results on RealDashCam(RDC) dataset.} For a clear visualization, we only draw the retained correspondences after running RANSAC. 
Correct correspondences are drawn in green while incorrect ones are drawn in red. We determine the correctness by thresholding the epipolar distances of the ground-truth poses.
}
\label{fig:didi_vis_result}
\vspace{-20pt}
\end{figure*}

\begin{figure}
\begin{center}
    \includegraphics[width=1\linewidth]{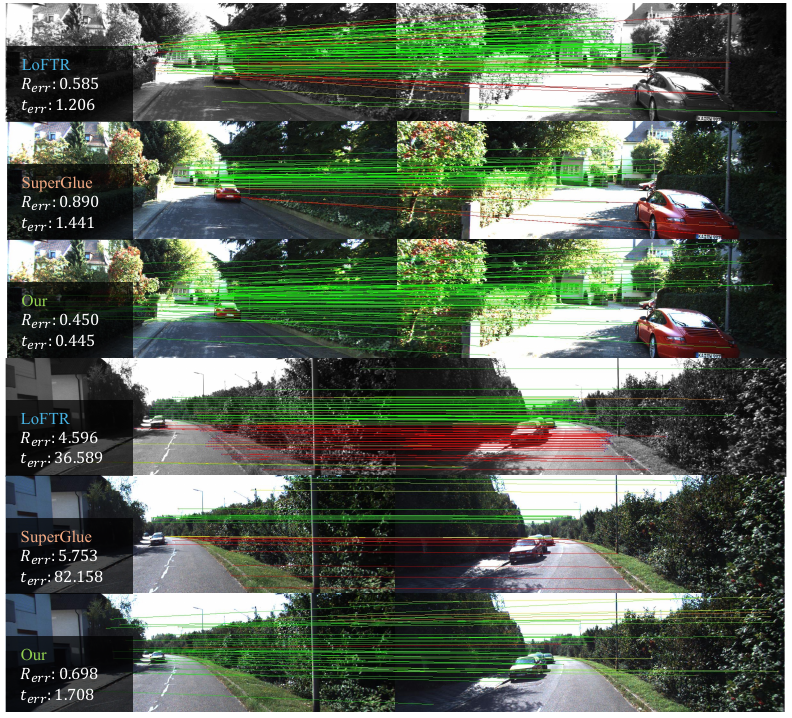}
\end{center}
\vspace{-10pt}
\caption{\textbf{Example result on KITTI dataset} In order to make a clear view, we simply draw the inlier correspondence of RANSAC. Correct correspondences are drawn in green while incorrect ones are drawn in red. In the KITTI dataset, the street view is clean and lacks moving objects, most of the test pairs perform the same.}
\label{fig:kitti_vis_result}
\vspace{-20pt}
\end{figure}

\paragraph{Implementation details} 
The pose regressor, matcher, and hypothesis score network are trained separately on the KITTI dataset using the Adam optimizer. The learning rate is set at $1\times 10^{-4}$, which undergoes annealing from $1\times 10^{-4}$ to $1\times 10^{-5}$.
{The regressor module uses a 5-layer CNN with output dimensions $64/256/512/1024/2048$, followed by average pooling and two MLPs for quaternion and transaction regression. The matcher module employs a 4-layer MLP encoder ($32/64/128/256$) and 9 self- and cross-attention layers. The pose regressor has three branches: descriptor (4-layer MLP with dimensions $512/256/128/64$), prior (4-layer MLP with dimensions $6/16/32/64$), and inlier (4-layer MLP with dimensions $64/64/128/128$). Features from these branches are concatenated and processed through a score net with dimensions $256/256/128/64/32/16/1$ to produce the score.}
The baseline models of matching were fine-tuned using the KITTI dataset, considering only correspondences with depth information. As NefSAC~\cite{cavalli2022nefsac} was also trained on the KITTI dataset, we utilized its pretrained weights. Due to the unavailability of the source code for MQNet~\cite{barath2022learning}, we re-implement it to the best of our ability.



\subsection{Performance}





\paragraph{Pose Estimation Performance}

{The qualitative comparison is shown in Fig.~\ref{fig:didi_vis_result} and Fig.~\ref{fig:kitti_vis_result}, while the quantitative results of the two datasets are presented in Tab.~\ref{tab:auc_result}}. 
\begin{itemize}
    \item The images within the RDC dataset exhibit a lower quality compared to the KITTI dataset {and the NuScenes dataset}, characterized by increased blurring and the presence of dynamic objects. Therefore, all methods perform much better on the KITTI dataset {and the NuScenes dataset} compared to the RDC dataset.
    \item {The detector-free methods~\cite{sun2021loftr, chen2022aspanformer, edstedt2023dkm, edstedt2024roma} perform better than detector-based methods for the ability to find dense correspondences.} These methods require resizing images to a specific resolution, which may result in the loss of fine details. Nevertheless, dense correspondences also introduce numerous erroneous matches, particularly on dynamic vehicles, significantly misleading hypotheses in sampling in camera pose estimation.
    \item {MQNet~\cite{barath2022learning} and NefSAC~\cite{cavalli2022nefsac} learn the certain regularities to select good hypotheses. MQNet~\cite{barath2022learning} is able to boost the performance of SuperGlue~\cite{sarlin2020superglue} on some of the datasets. However, its performance is limited to the correspondences given by SuperGlue~\cite{sarlin2020superglue} and may overfit KITTI.}
    \item {Our approach outperforms all of the datasets} by leveraging coarse camera poses, thus showcasing the effectiveness of our methodology.
\end{itemize}

\begin{figure}
\begin{center}
    \includegraphics[width=1\linewidth]{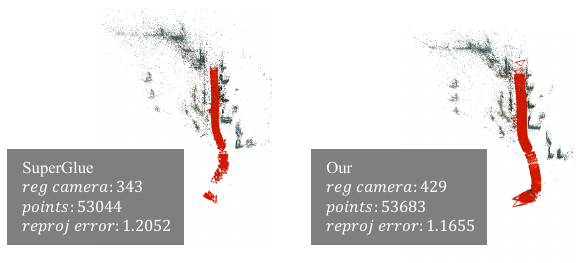}
\end{center}
\vspace{-10pt}
\caption{Reconstructed cameras and sparse points on the RDC dataset. {(One result from 10 trials)}}
\label{fig:sfm_sparse_result}
\vspace{-20pt}
\end{figure}

\paragraph{SfM Performance}
To further validate the performance of our matching method, we perform {SfM} on the RDC dataset. In this experimental setup, we utilize the correspondences estimated by our method as inputs to COLMAP to assess the quality of the reconstruction. Our evaluation of the SfM quality encompasses three key metrics: the number of registered cameras ("registered cameras"), the number of reconstructed points ("points"), and the projection error ("reproj. err.") measured in pixels. A registered camera refers to a camera whose pose has been successfully recovered and incorporated into the sparse model. We compare our method with SuperGlue~\cite{sarlin2020superglue}, presenting the quantitative results in Tab.~\ref{tab:sfm_sparse_result}, and visualizing the reconstructed trajectory and sparse points in Fig.~\ref{fig:sfm_sparse_result}.

Due to the inherent challenges posed by the RDC dataset, the outcomes of {SfM} applications may exhibit instability. Due to the random results, we execute COLMAP 10 times and report the average performance across the metrics.

Although our method may result in a lower number of sparse 3D points, the incorporation of camera motion {prior} allows for the identification of more reliable correspondences. 
Our approach is likely to produce fewer sparse 3D points because SuperGlue~\cite{sarlin2020superglue} might generate inaccurate correspondences for moving objects, leading to {noisy sparse points.}
As a result, our approach facilitates the successful registration of a greater number of cameras with reduced reprojection errors compared to SuperGlue~\cite{sarlin2020superglue}.

\begin{table}
\begin{center}
\caption{Quantitative results of SfM on the RDC dataset {(10 trials averaged)}}

\label{tab:sfm_sparse_result}
\resizebox{0.99\linewidth}{!}{
    \begin{tabular}{lccc}
    \toprule
    method & reg. camera & point &  reproj. err.(px) \\
    \midrule
    SuperGlue~\cite{sarlin2020superglue} & 341.1 & \textbf{56600.1} & 1.1939 \\
    Ours & \textbf{424.7} & 52269.9 & \textbf{1.1721} \\
    \bottomrule
    \end{tabular}}
    \vspace{-10pt}
\end{center}
\end{table}

\subsection{Ablation Analysis}
{To validate our design, we conduct ablation studies on the KITTI dataset for relative pose estimation. The quantitative results of our ablation studies are shown in Tab.~\ref{tab:ablation_result}.
In this table, "pos. reg." denotes the pose regressor, "SG" refers to SuperGlue~\cite{sarlin2020superglue}, "epi. enc." signifies the epipolar encoding, and "prior est. (samp.)" represents the proposed pose estimation method utilizing our sampling implementation.}

Ablation model 0 refers to the pose regression component, while ablation model 1 represents the SuperGlue~\cite{sarlin2020superglue} matcher with RANSAC pose estimation. Ablation model 2 indicates our matcher integrated with epipolar encoding and RANSAC pose estimation. Ablation model 3 denotes the SuperGlue~\cite{sarlin2020superglue} matcher coupled with our pose estimator incorporating pose prior information. Finally, ablation model 4 encompasses our complete model.


\begin{table}
\begin{center}
\caption{{Ablation studies on the KITTI dataset}}
\label{tab:ablation_result}
\resizebox{0.99\linewidth}{!}{
    \begin{tabular}{clccc}
    \toprule
    \multirow{2}*{ID} & \multirow{2}*{Method} & \multicolumn{3}{c}{AUC} \\
    \cmidrule(lr){3-5}
    & & $5^{\circ}$ & $10^{\circ}$ & $20^{\circ}$ \\
    \midrule
    0 & {pos. reg.}              & {0.7142} & {0.8469} & {0.9210} \\
    1 & {SG + RANSAC}            & {0.7454} & {0.8654} & {0.9303} \\
    2 & {SG + epi. enc.+RANSAC}  & {0.7547} & {0.8693} & {0.9318} \\
    3 & {SG + prior est.(samp.)}      & {0.7745} & {0.8714} & {0.9242} \\
    4 & {SG + epi. enc.+ prior est.(samp.)} & {\textbf{0.7998}}   &  {\textbf{0.8906}}   & {\textbf{0.9404}} \\
    \bottomrule
    \end{tabular}}
    \end{center}
    \vspace{-10pt}
\end{table}

\begin{itemize}
    \item {Comparing Ablation Model 0 with Ablation Model 1 demonstrates that the prior poses predicted by our pose regressor achieve high accuracy, comparable to SuperGlue~\cite{sarlin2020superglue}. his highlights the effectiveness of our pose regressor in capturing robust camera motion priors.}
    \item {When comparing Ablation Model 2 with Ablation Model 1, it is evident that epipolar encoding leads to improved performance across AUC metrics. The inclusion of epipolar line encoding enhances the matcher’s ability to identify correspondences of higher quality.}
    \item  {When comparing Ablation Model 3 with Ablation Model 1, it is evident that our method effectively samples robust epipolar geometries from the correspondences provided by SuperGlue~\cite{sarlin2020superglue}. In contrast to traditional RANSAC, which faces difficulties with hypothesis selection, our approach, supported by pose prior input, successfully preserves the correct hypotheses.}
    \item {Ablation Model 4 represents our comprehensive approach, integrating prior pose information for both correspondence estimation and pose estimation. As a result, it achieves superior performance in pose estimation.}
\end{itemize}

\subsection{Runtime}

\begin{table}
\centering
\caption{{Infer time analysis}}
\label{tab:time_result}
\resizebox{0.8\linewidth}{!}{
    \begin{tabular}{cllc}
    \toprule
    ID & Method & Type & Avg. time(s)\\
    \midrule
    0 & {pos. reg.}             & {Regressor} & {0.060} \\
    \midrule
    1 & {SuperGlue~\cite{sarlin2020superglue}}   & {Matcher} & {0.034} \\
    2 & {LoFTR~\cite{sun2021loftr}}              & {Matcher} & {0.051} \\
    3 & {ASpanFormer~\cite{chen2022aspanformer}} & {Matcher} & {0.083} \\
    4 & {DKM~\cite{edstedt2023dkm}}              & {Matcher} & {0.431}\\
    5 & {RoMa~\cite{edstedt2024roma} }           & {Matcher} & {0.384}\\
    6 & {SG + epi. enc. }                        & {Matcher} & {0.032} \\
    \midrule
    7 & {NeFSAC~\cite{cavalli2022nefsac}}         & {Estimator} & {2.311} \\
    8 & {samp.       }                            & {Estimator} &  {0.645}\\
    9 & {MQNet(samp.)~\cite{barath2022learning}} & {Estimator} & {0.654} \\ 
    10 & {prior est.(samp.)}                      & {Estimator} & {0.652} \\
    \midrule
    11 & {Our Full Pipeline} & {Pipeline} & {0.745} \\
    \bottomrule
    \end{tabular}}
    \vspace{-15pt}
\end{table}

Quantitative results are shown in Tab.~\ref{tab:time_result} {and the notation consistent with that in Tab.~\ref{tab:ablation_result}.}
{The inference time experiment was conducted on the KITTI dataset with a maximum iteration limit set at 1000.}
The detector-free matching method typically exhibits lower efficiency compared to detector-based matching methods due to its generation of dense correspondences for nearly all pixels. 
 {In the estimation process, the primary limitation is the sampling procedure. However, this procedure is highly optimized in OpenCV RANSAC.}



\section{Conclusion}
\label{sec:conclusion}


In this paper, we present a novel image matching framework for dashcam images.
We have observed a robust reviewed{motion prior} inherent in dashcams, which proves advantageous for the learning of correspondence and filtering outliers. Consequently, we employed a pose regression module to regress the motion {prior} of the camera, encoding them via soft epipolar constraint into the matcher. Simultaneously, we applied this methodology within random sample consensuses to assess their quality, thereby achieving precise pose estimation.
The experimental results validate that our method outperforms all the existing methods and supports all claims made in this paper. We believe that our framework will benefit the production and updating of high-definition map, as well as improve subsequent geo-information analysis tasks.



\vspace{-10pt}

\bibliography{glorified,new}
\bibliographystyle{IEEEtran}

\end{document}